\documentclass[10pt,letterpaper]{article}

\usepackage{cogsci}

\cogscifinalcopy %

\usepackage{graphicx} 
\usepackage{tabularx}
\usepackage[svgnames]{xcolor} 
\usepackage{amsmath}

\usepackage{pslatex}
\usepackage{apacite}
\usepackage{float} %

\title{Decoding Emotions in Abstract Art: Cognitive Plausibility of CLIP in Recognizing Color-Emotion Associations}
 
\author{{\large \bf Hanna-Sophia Widhoelzl  (hannasophia.widhoelzl@gmail.com)} \\
  Institute for Interdisciplinary Studies, University of Amsterdam \\
  Amsterdam, the Netherlands
  \AND {\large \bf Ece Takmaz (e.k.takmaz@uu.nl)} \\
  Department of Information and Computing Sciences, Utrecht University \\
  Utrecht, the Netherlands}

\begin{document}

\maketitle

\newcommand{\ece}[1]{\textcolor{orange}{[E: #1]}}
\newcommand{\hanna}[1]{\textcolor{purple}{[H: #1]}}
\newcommand{\hannahanna}[1]{\textcolor{blue}{[HH: #1]}}

\begin{abstract}
This study investigates the cognitive plausibility of a pretrained multimodal model, CLIP, in recognizing emotions evoked by abstract visual art. We employ a dataset comprising images with associated emotion labels and textual rationales of these labels provided by human annotators. We perform linguistic analyses of rationales, zero-shot emotion classification of images and rationales, apply similarity-based prediction of emotion, and investigate color-emotion associations. The relatively low, yet above baseline, accuracy in recognizing emotion for abstract images and rationales suggests that CLIP decodes emotional complexities in a manner not well aligned with human cognitive processes. Furthermore, we explore color-emotion interactions in images and rationales. Expected color-emotion associations, such as red relating to anger, are identified in images and texts annotated with emotion labels by both humans and CLIP, with the latter showing even stronger interactions. Our results highlight the disparity between human processing and machine processing when connecting image features and emotions.

\end{abstract}

\section{Introduction}

Abstract art's subjective and open-ended nature evokes intricate emotional responses in viewers. Its intrinsic ambiguity in both perception and interpretation has intrigued artists and researchers alike: abstract art presents an idiosyncratic challenge to our understanding of visual communication. Understanding emotions in abstract art is a complex aspect of human cognition, affected by individuals' education, cultural background, and social context. Previous research showed that features such as strokes, shapes, and visual harmony influence someone’s sentiment \cite{ko2016discovering, lu2012shape, sartori2015affective, zhao2014exploring}. Most studies focused on the interaction between color and emotion. While color-emotion associations vary with cultural factors and subjective interpretations, psychological studies identified universal, yet complex trends: Brighter colors such as yellow are associated with happiness, while red may signify anger. Blue is commonly linked to sadness, black to fear, and grey to boredom and depressive moods. Brown and green are often thought to evoke disgust \cite{hemphill1996note, liu2022colour, sutton2016color}.

A novel frontier in this domain is the intersection of artificial intelligence (AI) and human cognition. Recent advancements in Natural Language Processing (NLP) and Computer Vision (CV) gave rise to pretrained models such as CLIP (Contrastive Language-Image Pretraining) developed by OpenAI \cite{radford2021learning}. CLIP excels at language understanding and zero-shot image classification across diverse content types including realistic photographs, cartoons, and diagrams.\footnote{In a zero-shot classification setup, a model identifies and categorizes content not encountered during training. This tests the model's capacity to generalize its understanding to novel data.} CLIP was later integrated into various models that leverage its cross-modal understanding for downstream tasks \cite{agarwal2021evaluating}, such as 
generating images from textual prompts \cite{ramesh2022hierarchical}, describing images with natural language \cite{pmlr-v202-li23q, berrios2023towards}, answering questions related to images \cite{shen2022how}, and emotion recognition in naturalistic images \cite{bondielli2021leveraging}. The cognitive plausibility of such models, and the degree to which their behavior mirrors human cognitive processes, is a growing area of interest. The assumption is that these models capture key elements of human-like understanding, aiding in explaining aspects of human cognition and behavior \cite{kennedy2009cognitive, cogplausibility}. Uncovering potential disparities and shortcomings in the models' performance in emotion recognition in abstract art compared to human expectations contributes to creating more accurate artificial models of human cognitive processes. For instance, \citeA{phillips2015utility} argued that incorporating more cognitive plausible assumptions into computational models of language acquisition improved the model's performance and explanatory power. 

This research leverages CLIP to explore the potential of current state-of-the-art vision and language models in elucidating the intricate cognitive processes by which abstract art elicits emotions. We investigate whether the representations generated by CLIP can be effective in predicting these emotional responses, assessing cognitive plausibility by evaluating model outputs. We employ the FeelingBlue dataset, a multimodal corpus designed to scrutinize the emotional implications of color in language and art \cite{ananthram2023feelingblue}. The dataset encompasses abstract art images and textual justifications for the emotions assigned to those images. We investigate factors affecting CLIP's emotion recognition abilities, such as model size and types of prompts, concentrating on color-emotion associations. 

Our hypotheses are: (1) CLIP will demonstrate accuracy above chance level (five emotions, 20\%) in predicting emotions for abstract art images and textual justifications. (2) Using a similarity-based encoding approach with CLIP embeddings to approximate the emotion labels of novel stimuli, we expect to see improved emotion recognition in images, providing insights into the model's nuanced understanding of emotions. (3) Analyzing color-emotion interactions in images and textual justifications provided by annotators rationalizing their emotion allocation is expected to corroborate existing literature on color-emotion associations.

\section{Related Work}

Sentiment analysis of textual content has received extensive attention in the realm of NLP 
\cite{dang2020sentiment, nasukawa2003sentiment, solangi2018review}. However, the exploration of affective properties in images is relatively limited. Traditional CV generally focuses on predicting emotions from images of facial expressions, body postures, and gestures. In contrast, this study aims to understand emotions evoked by conceptual scenes, making it a distinct task. While both textual and visual content can be inherently implicit and ambiguous, the abstract nature of utilized images may render emotion classification more complex. Nevertheless, studies reported that images can potentially elicit stronger emotional impacts on humans than text \cite{casas2019images}. Consequently, the significance of training multimodal models to reach a better representation of human behavior and cognition is increasingly recognized.

Considering CLIP, a notable study by \citeA{bondielli2021leveraging} delved into the model’s behavior in emotion classification. CLIP exhibited promising results for recognizing emotions in naturalistic images under zero-shot settings. Fine-tuning CLIP improved performance substantially, emphasizing CLIP's capacity to effectively learn from visual and textual cues. This research contributes to understanding CLIP's adaptability to emotion-related tasks and its ability to capture affective aspects of visual content.

There is limited research focusing on AI models' cognitive plausibility for emotion recognition in abstract art. \citeA{yanulevskaya2008emotional} confirmed the potential of Support Vector Machines in perceiving emotions elicited through art by using a scene categorization system, while emphasizing the impact of painting techniques and color on emotion distinction. Their dataset depicted objects, people, and landscapes rather than consisting exclusively of fully abstract pieces. Thus, our study builds upon existing literature to explore the challenges and potential advancements of models in mirroring human emotion perception in abstract images. 

\section{Dataset}

\begin{figure*}[ht] 
\centering
\includegraphics[width=0.9\textwidth]{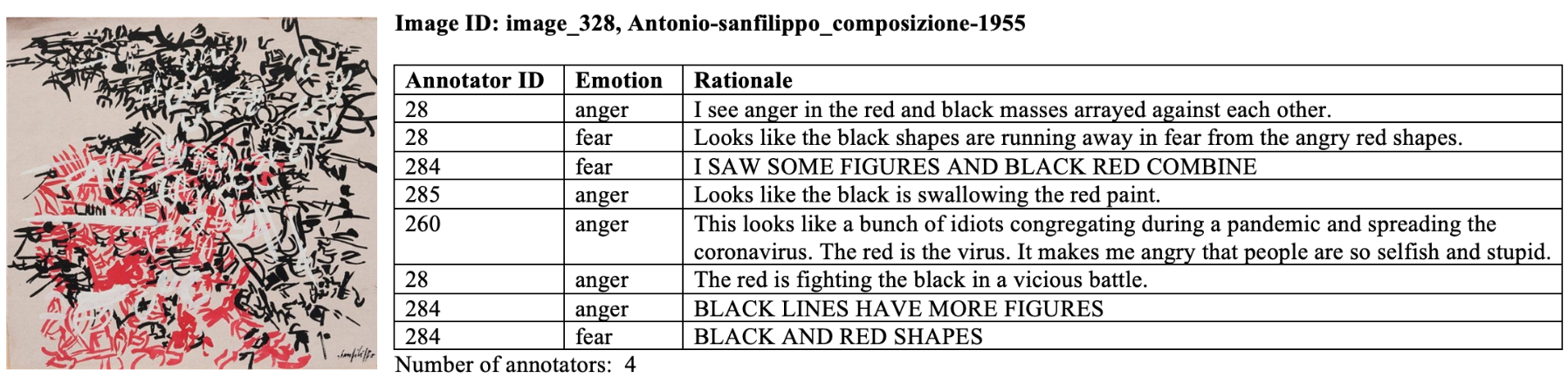}
\caption{Emotion allocation for `Composizione' by Antonio Sanfilippo (1955), from FeelingBlue \cite{ananthram2023feelingblue}.} 
\label{fig:1}
\end{figure*}

The FeelingBlue corpus \cite{ananthram2023feelingblue} explores the implications of changes in the colors of abstract paintings on emotional connotations in various contexts, mediated by visual elements such as lines, strokes, shapes, and language. It comprises images sourced from WikiArt \cite{mohammad2018wikiart} and DeviantArt \cite{sartori2015s}. Identifiable objects (e.g., flowers, statues) were excluded to eliminate confounding factors on perceived emotions. The corpus contains five emotion subsets: anger, disgust, fear, happiness, and sadness. FeelingBlue consists of 19,788 randomly generated 4-tuples of abstract art images. Annotators systematically ranked the images within the tuples based on a given emotion label according to Best-Worst Scaling (BWS) \cite{flynn2014best}. BWS involves annotators selecting the `best' and `worst' options from a set, revealing relative preferences across the dataset. In FeelingBlue, annotators were instructed to select images that most and least evoked the stated emotion within a tuple. This might lead to overlapping emotion subsets, where one image may be labeled with different emotions when presented in different contexts (Figure~\ref{fig:1}). Subsequently, annotators provided textual justifications, referred to as `rationales', for their choices of the `least' and `most' ranked image for evoking the specified emotion by describing salient visual features.

\section{Methods}

\subsection{Model}

CLIP \cite{radford2021learning} was trained on 400 million image-text pairs derived from the web. Its architecture involves an image encoder and a text encoder to map images and text into a shared vector space. During pretraining, CLIP’s contrastive training objective aims to increase the alignment score between a given image from the dataset and the text associated with this image (its caption). CLIP's ability to associate various visual concepts with corresponding captions generalizes to robust zero-shot performance in a plethora of vision and language tasks, including image and text emotion classification. Thus, CLIP seems a potentially valuable tool for exploring emotion recognition in abstract images. We employ two pretrained versions of CLIP's visual encoder: ViT-B/32 and ViT-L/14, based on the Vision Transformer architecture \cite{dosovitskiy2021an}. ViT-L/14, while significantly larger and demanding greater computational resources, offers potentially superior performance to B/32. To assess CLIP’s performance in subsequent analyses, we compute a score that measures the scaled cosine similarity between the representation of the input (images or rationales; $i$) and emotion labels as captions ($c$) in the form of vectors, which is called CLIPScore ($ \text{CLIPScore}(i, c) = 2.5 * \max(\cos(i, c), 0) $) \cite{hessel2021clipscore}. CLIPScore has been proposed as a metric to evaluate model-generated image captions, outputting scores well-aligned with human judgments.

\subsection{Analyses}

\subsubsection{Data Preprocessing}

We preprocess the FeelingBlue corpus to generate an image-emotion dictionary that facilitates further analyses and interpretations of results. In the original dataset, individual images are evaluated by multiple annotators on multiple emotions. For instance, image\_328 is ranked as evoking the most `anger’ five times and the most `fear’ three times within different 4-tuple image groups. Four different annotators assessed image\_328 (Figure~\ref{fig:1}). Participants could encounter the same images multiple times in distinct image groups or under different emotion label queries. We assign the emotion mentioned most frequently, in this case, `anger', to each image to create the aforementioned dictionary featuring 901 image-emotion pairs.\footnote{We picked the most frequently mentioned emotion as a simplification to facilitate further analyses; however, we acknowledge the complexity of emotional responses to images, which may simultaneously evoke various emotions or differ among viewers.} While the corpus initially contains 934 images, 33 images could not be assigned an emotion as they do not rank highest within any given tuple, prompting their removal from our dataset. Among the remaining images, 30.3\% are labeled as `happiness', 9.5\% as `anger', 8.8\% as `sadness', 16.0\% as `fear', and 35.4\% as `disgust'. The rationales are preprocessed using NLTK \cite{loper2002nltk}. All text is converted to lowercase. British spelling is converted to American spelling.

\subsubsection{Linguistic Features of Rationales}

We analyze the most common adjectives describing images linked to specific emotions mentioned in the rationales, focusing on color-related terms (9,912 rationales in total). Rationales for images rated as evoking anger, sadness, happiness, fear, or disgust are accordingly grouped by these emotions.

\subsubsection{Zero-Shot Emotion Classification}

We test  CLIP’s zero-shot ability in two setups using ViT-B/32 and ViT-L/14: (1) rationale to emotion label, (2) image to emotion label. In the former, rationales serve as textual input, and CLIP is presented with various emotion prompts. Thus, we compare text to text here by considering their similarity in CLIP's textual representation space, leveraging the model's robust understanding of textual semantics. In the latter, we compare each image to the emotion prompts. Thus, we compare text to image here, mirroring CLIP's contrastive learning objective.

As reported by \citeA{bondielli2021leveraging}, CLIP is sensitive to the wording of its prompts describing the emotion labels. We reuse their best-performing prompt structure, referred to as `standard prompts’, which puts the emotion label in context. For control purposes, we include `single-word prompts’. Finally, we create `art-tailored prompts’ to accommodate the nature of the stimuli by tailoring prompts to align with the specific themes of abstract art. We designed them through careful consideration of the distinctive features inherent in abstract art by stating the emotion label indirectly through artistic expressions. Terms representing the emotion labels are chosen based on the NRC Word-Emotion Association Lexicon (EmoLex) \cite{mohammad2013crowdsourcing}. EmoLex offers a list of English words associated with basic emotions identified through crowd sourcing, such as `melancholy’ linked to `sadness’. This results in three sets of prompt structures with five different prompts, each representing one emotion. Firstly, the `Standard' prompt: \{a sentence\textbar{}an image\} that elicits \{anger\textbar{}happiness\textbar{}sadness\textbar{}fear\textbar{}disgust\}. Secondly, the `Single-Word' prompt: \{anger\textbar{}happiness\textbar{}sadness\textbar{}fear\textbar{}disgust\}. Lastly, the `Art-Tailored' prompt: \{a sentence\textbar{}an image\} that evokes a sense of fiery intensity \textit{(anger)}; \{a sentence\textbar{}an image\} that captures a burst of vibrant joy \textit{(happiness)}; \{a sentence\textbar{}an image\} that conveys a profound sense of melancholy \textit{(sadness)}; \{a sentence\textbar{}an image\} that instils a feeling of eerie unease \textit{(fear)}; \{a sentence\textbar{}an image\} that evokes a sense of unsettling chaos \textit{(disgust)}. 

The prompt resulting in the highest CLIPScore when matched with a rationale or image is assumed to have the best fit with the input and is thus used to assign an emotion label. For example, if the prompt representing ‘happiness’ achieves the highest CLIPScore when matching the rationale `the sun is bright' with a set of prompts, the rationale is predicted to correspond to `happiness'. These predicted labels are subsequently compared to the emotion labels assigned by human annotators to calculate CLIP's accuracy.

\subsubsection{Similarity-Based Emotion Prediction}
We adopt a similarity-based encoding approach to further investigate the extent to which CLIP mirrors human emotion perception tendencies. The approach is inspired by \citeA{anderson2016representational}, who reported comparable accuracy to regression-based methods while obviating the need for model fitting when predicting fMRI signals for novel stimuli. The authors calculate the weighted average of the signals induced by a neighboring set of stimuli in the representational space to approximate the signals for a novel stimulus \cite{anderson2016representational}. In our work, we expect that images evoking similar emotions have similar representations in the shared feature space. Hence, we assess if images with similar representations by CLIP help predict the emotion label of the target stimulus. We divide the dataset into 80\% training set and 20\% test set. Emotion labels in the training set are transformed into one-hot vectors. We utilize the ten most similar training images for each test image by using cosine similarity between CLIP-encoded images. Using the ten most similar images is a trade-off between having a rich pool of samples to base the predictions on and computational efficiency. Subsequently, we calculate the weighted averages of the ten most similar images' emotion labels for each test image to predict its emotion label. 

\subsubsection{Color-Emotion Interaction}
Since literature suggests an apparent interaction between emotions and colors, we assess how color words mentioned in the rationales relate to emotions by probing for the following words that we identified through the linguistic features analysis: `black’, `dark’, `red’, `bright’, `yellow’, `white’, `blue’, `pink’, `orange’, `colorful, `grey’, `green’, `brown’, and `purple’. Firstly, we plot these words based on the emotions assigned to rationales by human annotators. We then examine the distribution of color terms in rationales based on CLIP’s predicted emotion labels. Due to computational constraints, we focused on ViT-B/32. We also analyze the interaction between emotions and the colors in images. Color distributions per human-annotated emotion are compared to CLIP ViT-B/32's emotion classification. To facilitate our analyses, we identify the most dominant color of images using ColorThief.\footnote{https://lokeshdhakar.com/projects/color-thief/} We map the resulting RGB values to color terms through Webcolors\footnote{ https://webcolors.readthedocs.io/en/latest/} (CSS Color Module Level 3). These color terms, such as `lime’, were then grouped into broader categories, such as `green’.

\section{Results \& Discussion}

\subsection{Images}

\subsubsection{Zero-Shot Emotion Classification}

CLIP’s accuracy for emotion recognition in images only slightly surpasses chance level under zero-shot settings (Table~\ref{table:1}), which is lower than anticipated, yet still supporting hypothesis (1). The highest accuracy is 29.12\% with ViT-L/14 and ‘art-tailored prompts’. The lowest accuracy of 11.56\% is observed with ViT-B/32 and ‘single-word’ prompts. CLIP’s performance is fairly sensitive to prompt wording: Providing the to-be-predicted emotion label in context leads to superior performance over stating the emotion in isolation \cite{bondielli2021leveraging}. Utilizing prompts tailored to art achieves the highest performance, suggesting that the context in prompts should relate to the nature of the images at hand. Yet ideally, models should attain a level of proficiency where they can be considered models of human cognition or used as tools for investigating human cognition, even when provided with single-word prompts. ViT-L/14 consistently outperforms B/32 as expected, given its larger model size and enhanced generalization abilities.

CLIP demonstrated lower accuracy in emotion recognition for abstract art than for naturalistic images \cite{bondielli2021leveraging}. Notably, the dataset used in the referenced study included images featuring individuals expressing emotions through facial expressions or body language. These additional cues might have contributed to higher accuracy. Additionally, differences in considered emotion classes and evaluation metrics require careful interpretation of the results when drawing direct comparisons with our findings. Nevertheless, this suggests that CLIP encodes emotional concepts in abstract images differently from naturalistic images.

\begin{table}[h]
\centering
\caption{Emotion recognition accuracy (\%) across data types, model versions, and prompt sets.}
\label{table:1}
\vskip 0.12in
\begin{tabular}{p{1.6cm}p{1.9cm}p{2cm}p{1.3cm}}
\hline
\textbf{Data Type} & \textbf{CLIP Model} & \textbf{Prompts} & \textbf{Accuracy} \\
\hline
Images & ViT-B/32 & Single-Word & 11.35 \\
 & & Standard & 11.56 \\
 & & Art-Tailored & \textbf{24.73} \\
 & ViT-L/14 & Single-Word & 14.12 \\
 & & Standard & 20.24 \\
 & & Art-Tailored & \textbf{29.12} \\
\hline
Rationales & ViT-B/32 & Single-Word & 34.79 \\
 & & Standard & \textbf{41.99} \\
 & & Art-Tailored & 26.85 \\
 & ViT-L/14 & Single-Word & \textbf{45.04} \\
 & & Standard & 44.71 \\
 & & Art-Tailored & 27.41 \\
\hline
\end{tabular}
\end{table}

\subsubsection{Similarity-Based Emotion Prediction}

Since CLIP ViT-L/14 did not achieve considerably higher accuracy than B/32 in the zero-shot experiments, we exclusively applied similarity-encoding to B/32 embeddings to predict images’ emotion labels. This resulted in an accuracy of 47.51\% on the test set, demonstrating superiority over random guessing and zero-shot performance, thereby corroborating hypothesis (2). This underscores the capability of CLIP's embeddings to decipher meaningful features associated with emotional content, as CLIP arrived at more correct predictions when considering similarly encoded images. Utilizing multiple similar samples enhances contextual awareness and potentially enables capturing more nuanced emotional subtleties that might be missed during zero-shot classification. This shows the importance of contextual awareness in emotion recognition of abstract art, highlighting the inherent strength of human cognition in a task that proves more challenging for machines when relying solely on visual cues, without access to the sociocultural environment \cite{mesquita2014emotions, mittal2021multimodal, yang2022emotion}.

Despite this improved accuracy, our findings suggest that CLIP's encoding of abstract images is not highly indicative of their emotional status. Alternatively, features of images eliciting the same emotion might vary considerably, thereby impeding CLIP’s emotion classification ability. Another possible, yet contradicting, hindrance is that features of images evoking different emotions might overlap, given the lack of concrete emotional cues. 

\subsubsection{Color-Emotion Interaction}
\begin{figure}[ht]
\begin{center}
\includegraphics[width=0.4\textwidth]{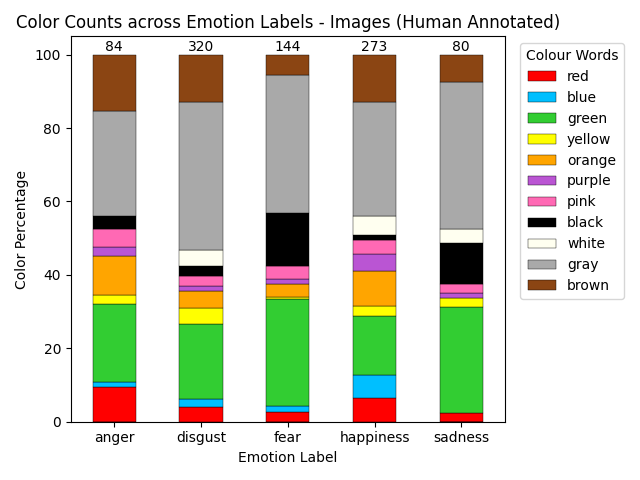}
\end{center}
\caption{ Color-emotion associations in human-annotated images with frequency counts per emotion category.} 
\label{fig:2}
\end{figure}

We explore the distribution of dominant colors of images per emotion label to identify color-emotion associations. Firstly, considering each image's dominant color and human-annotated emotion (Figure~\ref{fig:2}), we observe patterns aligning partially with our hypothesis (3). The dominant colors of `happiness' images are often orange, white, and yellow. Images of negative emotions, especially `fear’, include black. The `anger’ class has the highest proportion of red images, though not considerably more than `happiness’. Surprisingly, `happiness’ images are more often blue than ‘sadness’ images. Similarly, although `disgust’ images feature green or brown, we also see a large proportion of grey, as in all emotion categories. Instead, `fear’ images have the highest proportion of green. `Happiness’ and `disgust’ images show comparable amounts of brown.

\begin{figure}[ht]
\begin{center}
\includegraphics[width=0.4\textwidth]{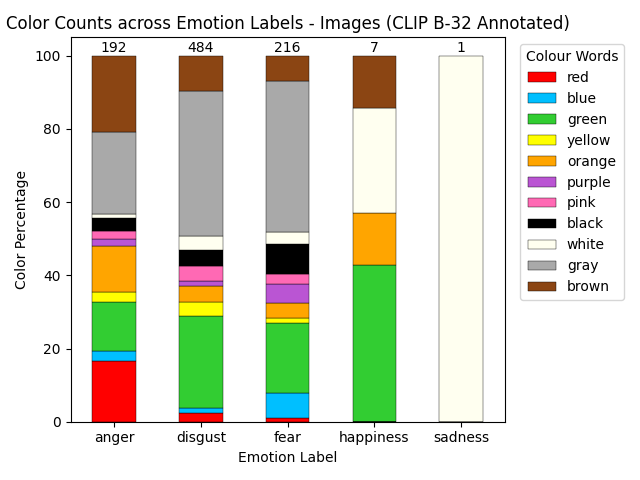}
\end{center}
\caption{Color-emotion associations in CLIP-annotated images with frequency counts per emotion category.} 
\label{fig:3}
\end{figure}

Different patterns of color-emotion distributions emerge in CLIP’s emotion classification for all included images, meaning both correctly and incorrectly labeled images, based on `art-tailored’ prompts (Figure~\ref{fig:3}). Firstly, `happiness’ images are frequently white. `Fear’, `disgust’, and `anger’ images have a considerable number of grey and black color. `Anger’ images have the highest proportion of red. While `happiness’ images feature the highest proportion of green, the `disgust’ category has the most green images among the negative emotions. Contrary to our hypotheses, the largest proportion of blue images is classified as `fear’ instead of `sadness’. `Anger’ images have the largest proportion of brown instead of `disgust’ images. Notably, CLIP categorizes only one image as `sadness’. The `sadness’-color distribution is therefore unrepresentative and should be disregarded. 

Interestingly, CLIP’s classification of images to emotions based on dominant colors aligns more closely with the theories from literature than the associations found between colors and human-annotated emotions. That CLIP’s emotion recognition has not achieved high accuracy, however, indicates that factors such as shapes, lines, textures, or overall compositions influenced the annotators' choices for emotion labeling. For CLIP, these factors might have proven more difficult to integrate into a holistic understanding of the images’ emotions.

\subsection{Rationales}

\subsubsection{Linguistic Features of Rationales}
Linguistic analyses of rationales reveal associations between emotions and descriptive language, including color-emotion interactions. Among images associated with happiness, commonly used adjectives are  `attractive', `bright', and `colorful'. These adjectives show limited overlap with those frequently found in annotations describing images associated with negative emotions, namely `red', `black', `dark', and `rough'.

\subsubsection{Zero-Shot Emotion Classification}

CLIP’s zero-shot emotion recognition performance in rationales is considerably higher than in images with an accuracy of 45.04\% with ViT-L/14 and `single-word’ prompts. Several factors might contribute to this. Firstly, textual justifications offer more explicit emotional cues, enabling CLIP to discern emotions more effectively compared to abstract images. CLIP may have learned linguistic patterns associated with certain emotions. Again, ViT-L/14 outperforms B/32 for all prompts. The lowest performance is attained with `art-tailored’ prompts, revealing an opposite pattern as found with images. This might be due to rationales providing context through concrete and abstract words. Thus, a concise prompt might be easier for CLIP to match with rationales. Abstract art images, instead, lack concrete concepts. CLIP might thus benefit from additional context, while humans might naturally derive emotionally meaningful interpretations from abstract art.

\subsubsection{Color-Emotion Interaction}
\begin{figure}[ht]
\begin{center}
\includegraphics[width=0.5\textwidth]{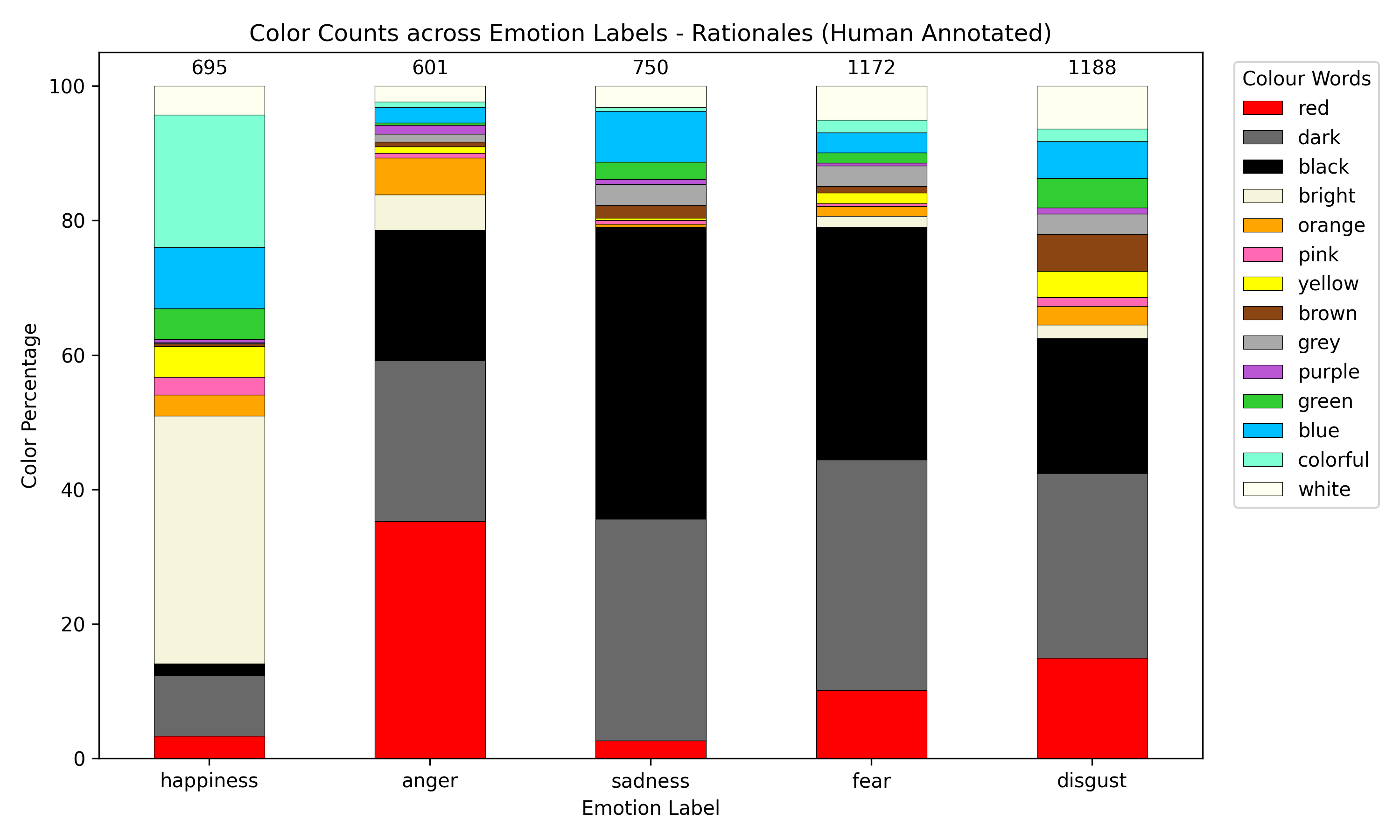}
\end{center}
\caption{Color-emotion associations in human-annotated rationales with frequency counts per emotion category.} 
\label{fig:4}
\end{figure}

Examining the distribution of color words mentioned in rationales based on human annotations reveals the following patterns (Figure~\ref{fig:4}): Words such as `bright’ and `colorful’ are commonly found in rationales describing `happy’ images. `Black’ and `dark’ are prevalent in rationales labeled with negative emotions. `Blue' is more frequent in `sadness' rationales than other negative emotions, while `red' is associated with `anger'. `Green' and `brown' appear more often in `disgust' rationales, yet these color-emotion associations are less pronounced. A substantial proportion of ‘fear’ rationales contain ‘black’, but the proportion of ‘black’ in ‘sadness’ rationales is even larger. These findings align with our initial hypotheses, yet highlight the complexity of color-emotion interactions.

\begin{figure}[ht]
\begin{center}
\includegraphics[width=0.5\textwidth]{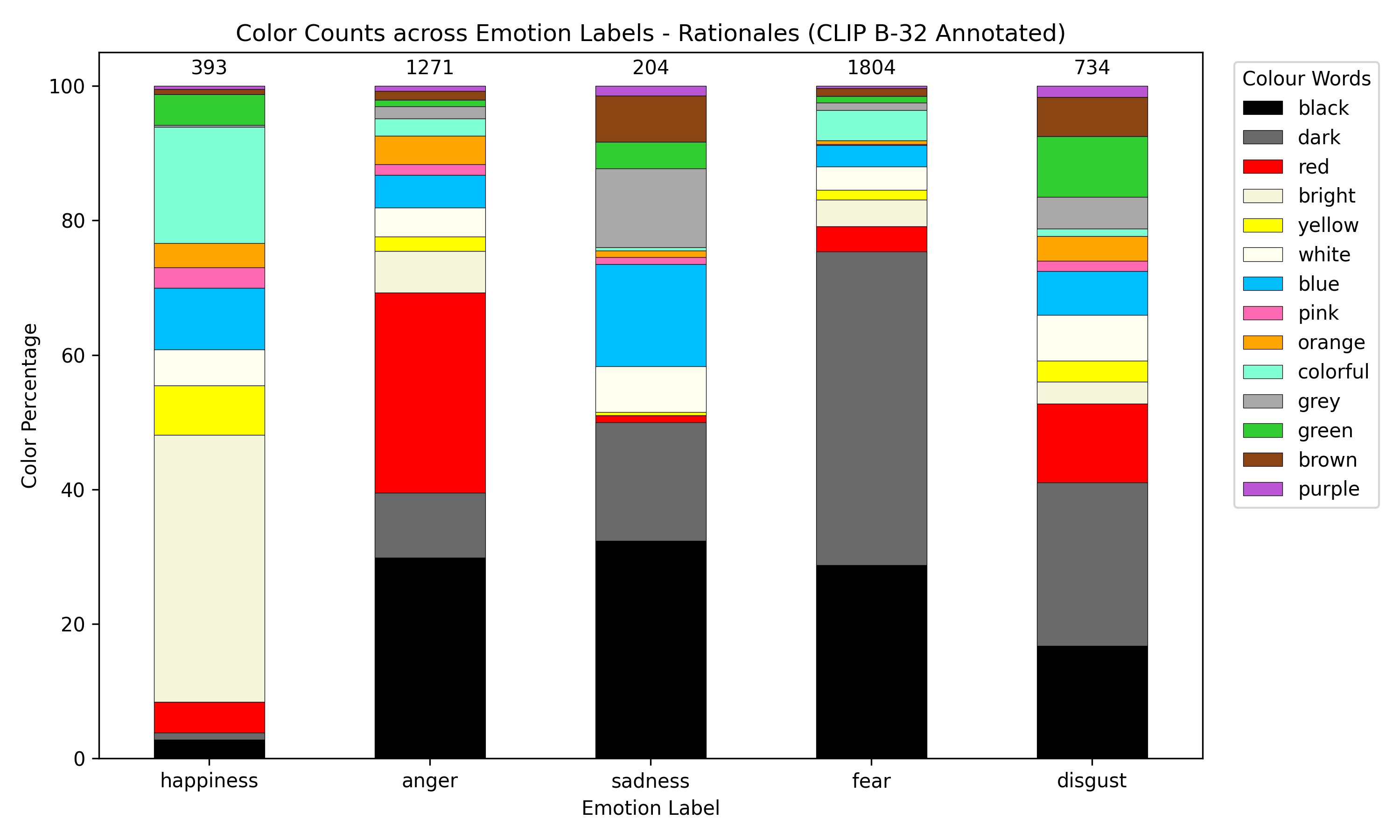}
\end{center}
\caption{Color-emotion associations in CLIP-annotated rationales with frequency counts per emotion category.} 
\label{fig:5}
\end{figure}

Results of the color word distributions in rationales according to CLIP’s classification can be seen in Figure~\ref{fig:5}. Again, `happiness’ rationales contain higher proportions of words such as `bright’ or `colorful’. Negative emotions are associated with `black’ and `dark’. The associations between `anger’ and  `red’, `sadness’ and `blue’, and `disgust’ and `green’ and `brown’ are more pronounced than in the human-annotated pattern. Overall, these patterns mostly support our hypothesis (3), which reflects the relatively high accuracy of rationale-emotion labeling with CLIP. 

CLIP's potential to recognize color-emotion associations in both images and text validates the relevance of color in emotional connotations. As found with images, CLIP’s categorization of rationales to emotions adheres to established color-emotion associations more strongly than human annotations. Hence, even for rationales mentioning explicit color words, other factors affect emotion perception in humans.

\section{General Discussion}

This study contributes to the growing body of research at the intersection of AI and art, paving the way for future exploration of human cognition and emotion through the lens of AI models. Exploring CLIP as a proxy model unveiled promising prospects for leveraging advanced vision and language models in decoding the emotional complexities of abstract art and textual justifications while reflecting on distinctive strengths of human cognition, such as context awareness. Overall, CLIP seems to exhibit difficulties bridging the affective gap between low-level features and abstract emotions, as there was no convincing evidence that CLIP latently encodes emotion elicited by images rigorously. This suggests potential limitations in its cognitive plausibility for tasks related to understanding and representing complex emotional states. That CLIP recognizes emotions in abstract art less accurately than in naturalistic images under zero-shot settings may be attributed to the inherent challenges of abstract art's subjective and ambiguous nature. Abstract art often lacks recognizable objects or scenes that may provide apparent emotional context, which may have hindered CLIP's ability to identify subtle emotional nuances and fine-grained details. Humans excel at capturing nuanced emotional expressions, leveraging a broader contextual understanding inherent in human cognition, such as personal experience, cultural influences, or recognition of familiar objects and scenes \cite{barrett2011context, hess2016impact}. These challenges faced by CLIP highlight a potential disparity between machine processing and human processing of emotions and abstract art, which necessitates further research from a cognitive modeling perspective.

Limitations include the subjective nature of abstract art, posing challenges for human annotators and CLIP. Social and cultural factors influence emotional judgments, hindering accurate insights into CLIP's performance as human annotators also demonstrated disagreement in emotion labeling. To acknowledge emotion's ambiguous nature, emotion classification should be operationalized as a multi-label task. The predominant focus on negative emotional states with a limited representation of positive emotions calls for future research to use emotion classes rooted in established psychological theories, such as \citeA{mikels2005emotional}'s model. This could offer a more cognitively meaningful coverage of emotions to enhance AI model's interpretability and relevance to human emotional experiences.

\section{Acknowledgments}

We thank Raquel Fernández and  Esam Ghaleb for their feedback on a preliminary version of this work, and the anonymous CogSci reviewers for their valuable comments on the paper. ET worked on this project while she was employed at the University of Amsterdam, where this research received funding from the European Research Council (ERC) under the European Union’s Horizon 2020 research and innovation program (grant agreement No. 819455).

\bibliographystyle{apacite}

\setlength{\bibleftmargin}{.125in}
\setlength{\bibindent}{-\bibleftmargin}

\bibliography{cogsci_full_paper_template}

\end{document}